%% file: main.tex
\documentclass[11pt]{article}

\usepackage[letterpaper,left=1.15in,right=1.05in,top=0.75in,bottom=0.8in]{geometry}
\usepackage{fontspec}
\newfontfamily\arabicfont[Script=Arabic,
  BoldFont=Amiri-Bold.ttf,
  ItalicFont=Amiri-Italic.ttf,
  BoldItalicFont=Amiri-BoldItalic.ttf
]{Amiri-Regular.ttf}
\newcommand{\textarabic}[1]{%
  \begingroup\arabicfont\beginR #1\endR\endgroup}
\usepackage{microtype}

\usepackage{amsmath,amssymb,amsthm}
\usepackage{graphicx}
\usepackage{booktabs}
\usepackage{tabularx}
\usepackage{placeins}
\usepackage{float}
\usepackage{natbib}
\usepackage{xcolor}
\definecolor{paperlinkblue}{RGB}{0,0,200}
\usepackage[
  colorlinks=true,
  linkcolor=paperlinkblue,
  citecolor=paperlinkblue,
  urlcolor=paperlinkblue
]{hyperref}
\usepackage[nameinlink,noabbrev]{cleveref}
\usepackage{fancyhdr}

\setcitestyle{authoryear,square,comma,aysep={,}}
\fancypagestyle{paper}{%
  \fancyhf{}
  \fancyfoot[C]{\small\thepage}
  
  }
\fancypagestyle{firstpage}{%
  \fancyhf{}
  \fancyfoot[L]{\small Preprint.}
  
  }
\graphicspath{{figures/}}

\theoremstyle{definition}

\theoremstyle{remark}

\makeatletter
\renewcommand{\maketitle}{%
  \thispagestyle{firstpage}%
  \vspace*{-1.2em}%
  \noindent\rule{\textwidth}{5pt}\par
  \vspace{1.25em}%
  {\raggedright\fontsize{19}{23}\selectfont\bfseries\@title\par}%
  \vspace{1.1em}%
  \noindent\rule{\textwidth}{0.8pt}\par
  \vspace{2.7em}%
  \begin{center}
    {\large\bfseries\@author\par}
  \end{center}
  \vspace{2.0em}%
}
\makeatother

\renewenvironment{abstract}{%
  \begin{center}
    {\large\bfseries Abstract\par}
    \vspace{0.8em}
    \begin{minipage}{0.84\textwidth}
      \small
}{%
    \end{minipage}
  \end{center}
  \vspace{0.9em}
}

\title{Detecting Hallucinations and Recovering Verified Answers in Arabic Islamic Question Answering}

\author{%
  Khaled Ziani\\[0.4em]
  \normalfont\normalsize Independent Researcher, Paris, France
}

\date{}

\begin{document}

\maketitle

\begin{abstract}
Large language models can generate fluent responses to Islamic questions while introducing factual errors that are difficult to identify. This paper presents our system for \textsc{HalluScoring 2026} Task 2.1, \textit{Islamic Hallucination Detection and Find the Truth}. The task requires a unified two-step prediction: determining whether an Arabic answer generated by an LLM is hallucinated and selecting the verified answer from six closely related candidate options. We use the Islamic knowledge dataset provided by the shared task, which contains 600 question--answer instances, including 341 hallucinated and 259 non-hallucinated answers. Our system is based on the fine-tuned \texttt{google/gemma-4-12B-it} model and uses deterministic decoding during inference. The generated outputs are normalized to extract the hallucination label and the selected option. The system achieves a Macro-F1 score of 0.928 and a label accuracy of 0.935 for hallucination detection, together with an option accuracy of 0.895 for answer selection. These results yield a combined score of 0.912, demonstrating strong performance across both stages of the task. The lower option-selection accuracy indicates that distinguishing the verified answer from plausible alternatives remains more challenging than detecting hallucinated responses.
\end{abstract}

\input{intro}
\input{data}
\input{exp}

\input{res}
\FloatBarrier
\input{error_analysis}
\input{conclusion}

\bibliographystyle{plainnat}
\bibliography{references}

\end{document}

%% file: intro.tex
\section{Introduction}
\label{sec:introduction}

Large language models (LLMs) can generate fluent, detailed, and persuasive
answers, but linguistic fluency does not guarantee factual accuracy. In
question answering, a model may invent a source, alter a quotation, confuse
entities or dates, provide an incorrect numerical value, or construct an
apparently coherent explanation around a false claim. Such errors are
particularly difficult to assess when the final conclusion is correct but part
of the supporting explanation is inaccurate. Reliable evaluation must
therefore examine the factual content of the entire response rather than only
its final answer.

Hallucination is commonly discussed in relation to both \emph{factuality} and
\emph{faithfulness}. A factuality error contradicts established knowledge or
introduces a claim that cannot be supported, whereas a faithfulness error
departs from the provided context, the user instruction, or the reasoning
presented in the response \citep{huang2025survey}. Although these two forms of
error often overlap, they may arise from different sources. Hallucinations can
result from incomplete, contradictory, or noisy pretraining data, biases in
instruction-tuning examples, the next-token prediction objective, failures to
retrieve relevant knowledge, or errors accumulated during multi-step reasoning
\citep{huang2025survey,alansari2025survey}. Hallucination detection is therefore
not merely a test of model confidence, since an incorrect statement may still
be produced fluently, consistently, and with high apparent certainty.

Existing hallucination-detection methods rely on several types of evidence.
Retrieval-based approaches compare generated claims with external documents or
structured knowledge sources. Uncertainty-based methods use token
probabilities, semantic entropy, internal model representations, or
disagreement across multiple generations. Supervised approaches instead learn
from manually or automatically annotated examples. Other methods employ
natural language inference, embedding similarity, self-consistency, or an LLM
as a judge \citep{alansari2025survey}. However, each signal has important
limitations. Retrieval-based methods depend on the coverage and reliability of
the retrieved evidence, supervised detectors may not generalize beyond their
annotation distribution, and self-consistency cannot guarantee correctness
when the model repeatedly generates the same false claim.

Benchmark design is consequently central to reliable hallucination evaluation.
\textsc{TruthfulQA} targets common misconceptions, \textsc{HaluEval} combines
synthetic examples with human annotation, \textsc{FELM} provides fine-grained
factuality judgments, and \textsc{RAGTruth} studies hallucinations in
retrieval-augmented generation. \textsc{Mu-SHROOM} extends span-level
hallucination detection to a multilingual setting. These resources cover
different domains, generation settings, and grounding conditions. Nevertheless,
many evaluations continue to reduce an entire response to a single binary
label. Although useful, such a label remains coarse: a completely fabricated
answer and an otherwise correct response containing one incorrect date receive
the same decision. Moreover, detecting that an answer is unreliable does not
necessarily demonstrate that a model knows the correct information.

These challenges are especially important for Arabic, where annotated
factuality resources remain comparatively limited and where methods developed
primarily for English cannot be assumed to transfer without adaptation
\citep{alansari2025survey}. Arabic morphology, lexical variation, spelling
variation, and alternative surface forms for named entities complicate the
comparison of generated claims with reference evidence. In knowledge-sensitive
domains, evaluation must also distinguish small but consequential differences
in legal terminology, numerical values, Qur'anic quotations, hadith
attributions, and source provenance.

Islamic knowledge represents a particularly sensitive setting because an
incorrect response may be presented together with apparently authoritative
religious evidence. In their evaluation of LLMs on Islamic inheritance-law
reasoning, \citet{bouchekif2025assessing} observed that some evaluated models,
particularly open-source models, fabricated Qur'anic verses or incorrectly
attributed non-existent formulations to the Qur'an when justifying their
answers. Such responses may appear credible because they imitate the form of
religious evidence while relying on false or misattributed content. This
highlights the need to evaluate both the final conclusion and the evidence used
to support it.

Recent Arabic benchmarks have begun to address different aspects of this
problem. \textsc{Halwasa} examines sentences generated from predefined Arabic
keywords, while \textsc{IslamicEval} evaluates Qur'anic and hadith-related
knowledge. \textsc{Aftina} studies retrieval-based mitigation for Islamic fatwa
generation. \textsc{AraHalluEval} covers Arabic question answering and
summarization, whereas \textsc{HalluScore} provides supporting evidence and
reviewed explanations for multi-domain questions. Arabic examples are also
included in multilingual resources such as \textsc{Mu-SHROOM} and
\textsc{HalluVerse25}. However, most existing settings primarily evaluate
hallucination detection or mitigation and do not explicitly assess whether a
model can recover the verified answer after identifying an unreliable
response.

\textsc{HalluTruthQA} addresses this distinction through a fine-grained Arabic
benchmark covering Islamic knowledge, history, science, and geography
\citep{bouchekif2026hallutruthqa}. Each instance includes a response-level
hallucination label, character-level error spans, a human-written explanation,
a hierarchical hallucination taxonomy, a verified reference answer, and six
candidate answers for factual verification. The benchmark therefore supports
the evaluation of several related but distinct capabilities, including
hallucination detection, error localization, explanation, and answer recovery.
Its results indicate that recognizing an unreliable response does not
necessarily imply that a model can locate the error or identify the correct
answer.

In this work, we present our system for \textsc{HalluScoring 2026} Task 2.1,
\textit{Islamic Hallucination Detection and Find the Truth}. The task follows a
unified two-step evaluation setting. Given an Arabic Islamic question and an
LLM-generated answer, the system first determines whether the answer is
\textsc{Hallucinated} or \textsc{Non-Hallucinated}. It then selects the
verified answer from six closely related candidate options. This formulation
jointly evaluates error recognition and factual recovery, without assuming
that successful hallucination detection automatically leads to the correct
answer.

Our system is based on \texttt{google/gemma-4-12B-it} and uses task-specific
fine-tuning for hallucination detection and answer selection. The training data
cover a range of factual errors involving entities, dates, numerical values,
legal rulings, quotations, source attributions, and unsupported claims. On the
official Islamic knowledge evaluation data, the system achieves a Macro-F1
score of \(0.928\) and a label accuracy of \(0.935\) for hallucination
detection, together with an option accuracy of \(0.895\) for factual answer
selection. The equally weighted combination of Macro-F1 and option accuracy
produces a final score of \(0.912\). The difference between the two scores
suggests that identifying the verified answer among plausible alternatives
remains more challenging than recognizing that a generated response contains
an error.

%% file: data.tex
\section{Training Data}
\label{sec:training_data}

We assembled separate Arabic training sets for hallucination detection and MCQ
answer selection. The detection corpus contains 43{,}400 question--answer pairs
drawn from Islamic jurisprudence, inheritance reasoning, Islamic knowledge,
Arabic and Islamic culture, and general knowledge. It covers both coarse factual
errors---for example, an incorrect ruling or answer---and localized errors in
names, dates, numerical values, Qur'anic quotations, and hadith attributions.
The selection corpus contains 15{,}100 questions with verified answers and
plausible distractors. Table~\ref{tab:training_data} gives the contribution of
each source.

\begin{table*}[t]
\centering
\footnotesize
\setlength{\tabcolsep}{4pt}
\begin{tabularx}{\textwidth}{@{}
  >{\raggedright\arraybackslash}p{2.8cm}
  >{\raggedright\arraybackslash}p{1.7cm}
  rr
  >{\raggedright\arraybackslash}X@{}}
\toprule
\textbf{Source} &
\textbf{Original format} &
\textbf{Detection} &
\textbf{MCQ selection} &
\textbf{Use} \\
\midrule

IslamWeb
& Fatwa QA
& 8{,}000
& --
& Original jurisprudential questions and answers. \\

QIAS 2025 \cite{bouchekif2025qias}
& MCQ
& 21{,}700
& 2{,}000
& Correct options provide positive answers; distractors provide plausible factual errors. \\

QIAS 2026 \cite{bouchekif2026qias}
& MCQ
& --
& 12{,}500
& Islamic inheritance. \\

MAWARITH \cite{bouchekif2026mawarith}
& Open QA
& 600
& --
& Inheritance problems requiring legal and numerical reasoning. \\

PalmX 2025 \cite{alwajih2025palmx}
& MCQ
& 2{,}500
& --
& Questions on Arabic and Islamic cultural knowledge. \\

IslamicFaithQA \cite{bhatia2026faithful}
& Open QA
& 2{,}000
& --
& Islamic knowledge questions with reference answers. \\

HalluTruthQA \cite{bouchekif2026hallutruthqa}
& Annotated QA/MCQ
& 600
& 600
& Factuality labels, verified answers, and five plausible distractors. \\

IslamQA
& Fatwa QA
& 5{,}000
& --
& Controlled perturbations of Qur'anic verses and hadith quotations. \\

Gemini-generated data
& Generated QA
& 3{,}000
& --
& Naturally generated correct and incorrect general-knowledge responses. \\

\midrule
\textbf{Total}
& --
& \textbf{55{,}900}
& \textbf{15{,}100}
& -- \\
\bottomrule
\end{tabularx}
\caption{Composition of the training data. The MCQ-selection examples are
reused from QIAS 2025 and HalluTruthQA and do not increase the size of the
detection corpus.}
\label{tab:training_data}
\end{table*}

\FloatBarrier

\subsection{Hallucination-Detection Instances}

For MCQ sources, we converted each question into a binary detection instance.
The candidate answer was the gold option for approximately half of the
questions and an original distractor for the remainder. We retained the dataset
distractors rather than generating new negatives: they are topically relevant,
grammatical, and deliberately written to be confused with the correct answer.
Given question $q_i$, gold option $o_i^{+}$, and distractor set
$\mathcal{O}_i^{-}$, we construct candidate answer $a_i$ as

\[
a_i =
\begin{cases}
o_i^{+},
  & y_i=\text{\textsc{Non-Hallucinated}}, \\[2mm]
o_i^{-}, \quad o_i^{-}\in\mathcal{O}_i^{-},
  & y_i=\text{\textsc{Hallucinated}}.
\end{cases}
\]

The open-ended sources serve a different purpose. IslamWeb,
IslamicFaithQA, MAWARITH, and HalluTruthQA broaden the corpus beyond
short MCQ answers and introduce longer explanations, legal reasoning, and
numerical inheritance calculations. From IslamQA, we created 5{,}000 controlled
examples by editing one or more words in Qur'anic verses or hadith quotations.
The edits include substitutions, deletions, and insertions chosen to preserve
the fluency of the surrounding answer while changing the quoted text or its
factual meaning. This produces localized errors that are easy to overlook when
the rest of the response is correct.

We also generated 3{,}000 Arabic answers to general-knowledge questions with
Gemini. These responses add model-generated errors that are less regular than
the controlled quotation perturbations and less constrained than MCQ
distractors. After construction, we balanced the corpus to contain approximately
equal numbers of hallucinated and non-hallucinated answers. A detection instance
is represented as

\[
x_i=(q_i,a_i), \qquad
y_i\in
\left\{
\text{\textsc{Hallucinated}},
\text{\textsc{Non-Hallucinated}}
\right\},
\]

where $q_i$ is the question, $a_i$ the candidate answer, and $y_i$ the
factuality label.

\subsection{MCQ Answer-Selection Instances}

The answer-selection set combines 2{,}000 questions from QIAS 2025 with 600
six-option questions from HalluTruthQA. Each instance contains a question, one
verified answer, and a set of plausible alternatives. We preserve the original
options because their semantic proximity makes selection more demanding than
choosing among randomly sampled negatives. This task is complementary to
binary detection: the detector decides whether a given answer is reliable,
whereas the selection model must recover the correct answer when several
candidates appear credible.

%% file: exp.tex
\section{Experiments}
\label{sec:experiments}

We conduct our experiments using the Islamic knowledge data released
for \textsc{HalluScoring 2026}\footnote{\url{https://huggingface.co/datasets/Bekhouche/HalluTruthQA-4K}}, Task 2.1: \textit{Islamic Hallucination
Detection and Find the Truth}. The dataset contains 600 Arabic
question--answer instances, including 341 hallucinated answers
(56.8\%) and 259 non-hallucinated answers (43.2\%). It also contains
393 manually annotated hallucination spans.

Task 2.1 follows a unified two-step evaluation setting. Given an Arabic
question and an LLM-generated answer, the system first determines
whether the generated answer is \textsc{Hallucinated} or
\textsc{Non-Hallucinated}. It then identifies the verified answer from
six closely related candidate options. The system consequently produces
a hallucination label and an answer-option prediction for each
evaluation instance.

For hallucination detection, we use Macro-F1 as the primary metric
because it assigns equal importance to the hallucinated and
non-hallucinated classes, independently of their frequencies. Given the
class-level F1-scores, Macro-F1 is computed as

\[
\text{Macro-F1}
=
\frac{
F1_{\textsc{Hallucinated}}
+
F1_{\textsc{Non-Hallucinated}}
}{2}.
\]

We also report label accuracy, which measures the proportion of
instances for which the predicted hallucination label matches the gold
label.

The second step evaluates whether the system selects the correct answer
from the six candidate options. It is measured using option accuracy:

\[
\text{Option Accuracy}
=
\frac{\text{Number of correctly selected options}}
{\text{Total number of evaluated instances}}.
\]

The official combined score gives equal weight to hallucination
detection and answer selection:

\[
\text{Combined Score}
=
0.5 \times \text{Macro-F1}
+
0.5 \times \text{Option Accuracy}.
\]

Thus, achieving a high final score requires the system to both
accurately assess the factual reliability of the generated answer and
identify the correct answer among the closely related alternatives.

Our system is based on \texttt{google/gemma-4-12B-it}. During inference,
we set the temperature to \(0.0\), top-\(p\) to \(1.0\), and the maximum
number of newly generated tokens to 2{,}048. The generated outputs were
normalized before scoring to extract the predicted hallucination label
and option identifier.

%% file: res.tex
Table~\ref{tab:main_results} compares our results with the zero-shot baselines
reported by \citet{bouchekif2026hallutruthqa} for the Islamic knowledge subset.
Our proposed approach achieves a label accuracy of \(0.935\), a Macro-F1 of
\(0.928\), and an option accuracy of \(0.895\). Their equally weighted average
gives a combined score of \(0.912\).

\begin{table}[H]
\centering
\footnotesize
\begin{tabular}{lcccc}
\toprule
\textbf{Model} &
\multicolumn{2}{c}{\textbf{Hallucination Detection}} &
\textbf{MCQ} &
\textbf{Combined} \\
\cmidrule(lr){2-3}
& \textbf{Label Acc.} & \textbf{Macro-F1} & \textbf{Score} &
\textbf{Score} \\
\midrule
Qwen3-32B      & 0.868 & 0.864 & 0.863 & 0.864 \\
Falcon-H1R-7B  & 0.863 & 0.857 & 0.853 & 0.855 \\
ALLaM-7B       & 0.862 & 0.860 & 0.823 & 0.842 \\
SILMA          & 0.737 & 0.735 & 0.641 & 0.688 \\
FANAR-9B$^{\dagger}$ & 0.908 & 0.904 & 0.883 & 0.894 \\
\midrule
\textbf{Proposed approach} & \textbf{0.935} & \textbf{0.928} &
\textbf{0.895} & \textbf{0.912} \\
\bottomrule
\end{tabular}
\caption{Comparison on the Islamic knowledge data. For the HalluTruthQA
baselines, the MCQ column reports LO-Score and the combined score is the
equally weighted average of Macro-F1 and LO-Score. For the proposed approach,
the MCQ column reports option accuracy and the combined score follows the
official HalluScoring 2026 formula. $^{\dagger}$FANAR-9B is evaluated in a
self-detection setting.}
\label{tab:main_results}
\end{table}

Among the independently evaluated zero-shot baselines, Qwen3-32B obtains the
highest label accuracy and Macro-F1, while FANAR-9B records the strongest
baseline scores overall in its self-detection setting. The proposed approach
outperforms both references across hallucination detection and MCQ selection.
This comparison should nevertheless be interpreted with care: the baseline
models were evaluated zero-shot, whereas our system was fine-tuned specifically
for the shared task.

%% file: error_analysis.tex
\section{Error Analysis}
\label{sec:error_analysis}

To complement the aggregate results, we manually examined representative
errors produced by the model. The analysis reveals several recurring failure
patterns involving answer validation, factual attribution, ambiguity in the
reference answers, and inconsistencies between hallucination detection and
multiple-choice verification.

\paragraph{False acceptance of plausible but incorrect answers.}
A first family of errors occurs when the model predicts that a
Fanar-generated answer is non-hallucinated although it contains an incorrect
entity or attribution. These errors are particularly difficult because the
generated answer is often fluent, relevant to the question, and formulated in
a style similar to a correct answer. The model therefore appears to evaluate
the semantic plausibility of the response rather than verify its precise
factual content.

For example, one question asks which prophet made the supplication
``\textarabic{على الله توكلنا ربنا افتح بيننا وبين قومنا بالحق وأنت خير
الفاتحين}''. The generated answer attributes the supplication to Prophet
Hūd, and the detector incorrectly accepts this answer as non-hallucinated.
However, the quotation appears in the account of Prophet Shu\textquotesingle ayb
in Qur'an 7:89. This example illustrates an entity-attribution error: the
response remains thematically compatible with prophetic narratives, but the
named prophet is incorrect. Similar errors may involve prophets, companions,
narrators, historical figures, surahs, or Qur'anic verse numbers.

\paragraph{Reference-sensitive and ambiguous cases.}
A second category includes examples for which the generated answer differs
from the annotated reference but remains factually defensible. For the
question asking which Qur'anic verse establishes the obligation of fasting,
the reference answer gives Qur'an 2:183:
``\textarabic{يا أيها الذين آمنوا كتب عليكم الصيام}''. Fanar instead returns
Qur'an 2:185, which contains
``\textarabic{فمن شهد منكم الشهر فليصمه}''. Although the former verse states
the prescription of fasting more explicitly, the latter also directly
commands fasting during Ramadan.

Consequently, accepting the generated answer in this case should not
necessarily be interpreted as a factual verification failure. Rather, the
example exposes the sensitivity of the evaluation to a single reference
answer when multiple passages can reasonably answer the question. Such cases
may artificially increase the number of apparent false negatives. They also
highlight the importance of distinguishing genuine hallucinations from
alternative valid answers during annotation and adjudication. Questions with
more than one acceptable interpretation should either allow multiple
references or be rewritten to request a more specific answer.

\paragraph{Correct detection followed by incorrect option selection.}
Another important error family concerns inconsistencies between the two
evaluated tasks. In some cases, the model correctly identifies the generated
answer as hallucinated but subsequently selects an option that reproduces the
same incorrect information.

For instance, in response to the question
``\textarabic{من آخر من مات من الصحابة؟}'', Fanar answers Anas ibn Mālik,
whereas the reference answer identifies Abū al-Tufayl \textquoteleft Amir ibn
Wāthila. The model correctly predicts that the generated response is
hallucinated. Nevertheless, during multiple-choice verification, it selects
the option stating that the answer is Anas ibn Mālik. Thus, the model
recognizes that the initial response is unreliable without recovering the
factual knowledge required to correct it.

This result confirms that hallucination detection and factual verification
are related but distinct capabilities. A detector may learn textual patterns
associated with unreliable answers, or recognize a mismatch between the
question and the generated response, without knowing which alternative is
correct. High detection performance therefore does not necessarily imply
strong factual knowledge.

\paragraph{Anchoring on the generated answer.}
The preceding example also suggests an anchoring effect. The incorrectly
selected option repeats the Fanar answer almost verbatim. The model may
therefore favor the candidate with the greatest lexical or semantic overlap
with the generated response, even after classifying that response as
hallucinated. This behavior is especially problematic when distractors are
constructed from common misconceptions or are written in a fluent and
authoritative style.

Such errors indicate that the option-selection component does not always use
the hallucination decision consistently. When the answer is classified as
hallucinated, an option that reproduces its central incorrect claim should
normally receive a lower score. Training with hard negative options that
closely resemble the generated answer may help reduce this copying bias.
A consistency objective between the detection and selection tasks could also
discourage predictions in which the model rejects an answer and then selects
an equivalent candidate.

\paragraph{Fine-grained knowledge confusion.}
More generally, many of the inspected errors involve closely related
religious entities or facts rather than entirely unrelated answers. Examples
include confusing two prophets, selecting a well-known companion instead of
the last surviving companion, or retrieving a relevant Qur'anic verse that
does not match the expected reference. These cases require precise
fine-grained knowledge and cannot be resolved through topic recognition
alone. The model often identifies the correct semantic area while failing to
retrieve the exact entity, attribution, date, or textual source.

Overall, the qualitative analysis identifies three broader limitations.
First, the model may mistake plausibility for factual correctness and accept a
fluent but incorrectly attributed answer. Second, a binary evaluation may
penalize valid alternatives when the question or reference answer is
underspecified. Third, the model may detect the presence of a hallucination
without possessing sufficient factual knowledge to select the correct answer.
These findings motivate more careful reference-answer adjudication,
contrastive training on closely related entities, and stronger consistency
between hallucination detection and factual verification.

%% file: conclusion.tex
\section{Conclusion}
\label{sec:conclusion}

We presented our system for \textsc{HalluScoring 2026} Task 2.1, which jointly
evaluates hallucination detection and factual answer selection for Arabic
Islamic question answering. Using separate task-specific adapters based on
\texttt{google/gemma-4-12B-it}, the system achieved a Macro-F1 of \(0.928\)
and a label accuracy of \(0.935\) for hallucination detection, together with
an option accuracy of \(0.895\). These results produced a combined score of
\(0.912\).

The results also show that detecting an unreliable response and recovering the
correct answer are not equivalent capabilities. The model can reject a
generated answer while still selecting an option that repeats the same error.
Our qualitative analysis further identified failures involving plausible but
incorrect attributions, confusion between closely related entities, anchoring
on the generated response, and questions for which more than one answer may be
factually defensible. These cases suggest that evaluation based on a single
reference answer can sometimes conflate genuine hallucinations with valid
alternatives.

Future work should therefore combine stronger factual supervision with more
careful adjudication of reference answers. Contrastive examples involving closely
related entities and hard distractors may improve fine-grained discrimination,
while an explicit consistency objective could reduce cases in which the model
rejects an answer and subsequently selects an equivalent option. Extending the
evaluation to error localization and correction would also provide a more
complete account of factual reliability than response-level classification
alone.